# Generative Modeling using the Sliced Wasserstein Distance

Ishan Deshpande
University of Illinois
Urbana-Champaign
ideshpa2@illinois.edu

Ziyu Zhang
Snap Inc.
Los Angeles
zzhang3@snap.com

Alexander Schwing
University of Illinois
Urbana-Champaign
aschwing@illinois.edu

## Abstract

*Generative Adversarial Nets (GANs) are very successful at modeling distributions from given samples, even in the high-dimensional case. However, their formulation is also known to be hard to optimize and often not stable. While this is particularly true for early GAN formulations, there has been significant empirically motivated and theoretically founded progress to improve stability, for instance, by using the Wasserstein distance rather than the Jenson-Shannon divergence. Here, we consider an alternative formulation for generative modeling based on random projections which, in its simplest form, results in a single objective rather than a saddle-point formulation. By augmenting this approach with a discriminator we improve its accuracy. We found our approach to be significantly more stable compared to even the improved Wasserstein GAN. Further, unlike the traditional GAN loss, the loss formulated in our method is a good measure of the actual distance between the distributions and, for the first time for GAN training, we are able to show estimates for the same.*

## 1. Introduction

Generative modeling is a topic of increasing importance. In contrast to discriminative approaches, where significant progress has been made in the last decades, generative models are still at their infancy. This is partly due to the fact that the output space considered when modeling a distribution over data is often significantly larger. Because of this large output space, classical generative models such as probabilistic semantic indexing [10], restricted Boltzmann machines [30], or latent Dirichlet allocation [4] have to sample from high-dimensional distributions, which is challenging.

Instead, in recent years, progress in generative modeling suggests to make use of the manifold assumption, *i.e.*, to sample from simple distributions and subsequently transform the sample via function approximators such as deep nets to yield the desired output. Variational autoencoders [14] and adversarial nets [8] are among the algorithms which follow this paradigm. Variational autoencoders are based on the principle of a variational lower bound which is maximized. Their probabilistic interpretation is appealing, but they are known to produce samples that are often overly smooth when considering images. In contrast, generative adversarial nets are often intuitively explained using a two-player game analogy. They are known to produce sharp examples, but, among others due to the saddle-point formulation inherent to two-player games, optimization is finicky, as justified by the many papers addressing this topic [1, 9, 31, 26, 27, 22, 20, 25, 13, 11, 19]. Among the most pressing issues are mode dropping, vanishing gradients, training instability, and sensitivity to parameter initialization.

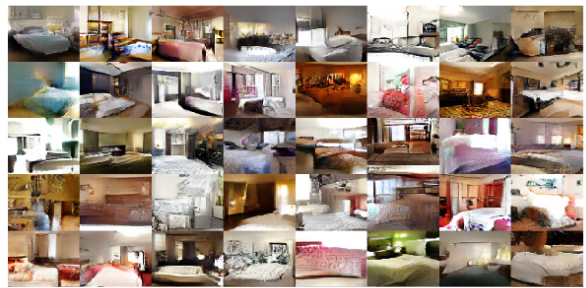

Figure 1. Generated samples from the LSUN bedrooms dataset.

To address some of the issues, most notably vanishing gradients, Arjovsky *et al*. [2] recently introduced a variant of GANs based on the Wasserstein distance rather than the classical Jensen-Shannon divergence. Their approach employs the Kantorovich-Rubinstein duality which results in a saddle-point objective, just like the original GAN framework. However, optimization of saddle-point objectives is challenging, particularly if neither of the directions is convex or concave. Hence, optimization of the Wasserstein GANs remains tricky as suggested by recent improvements [9]. Special techniques, *e.g*., [24, 23] are generally necessary but practically not used.

In this paper we improve the stability of Wasserstein GAN training by developing a mechanism based on random projections as opposed to using the Kantorovich-Rubinstein duality. Different from the duality-based approach, in its simplest form, we are able to formulate the optimization using a single minimization. To this end we utilize the "sliced Wasserstein distance," employed, *e.g.*, by Rabin *et al.* [28] for texture mixing. The sliced Wasserstein distance has since been studied and successfully applied to a variety of tasks such as color transfer [6] and image classification [15]. Recently, Kolouri *et al.* [16] also proposed a family of provably positive definite kernels based on the sliced Wasserstein distance and showed its efficacy on various pattern recognition tasks.

Beyond stability improvements the proposed formulation also enables a bound for attainable performance and simple extensions to address modeling in high-dimensional cases, *e.g.*, when considering images more complex than MNIST [18].

In experiments on the MNIST handwritten digit dataset [18], the Toronto face dataset [32], the CIFAR-10 dataset [17], the CelebA dataset [21], and the LSUN bedroom dataset [35] (see Fig. 1), we demonstrate that our approach is significantly more stable than conventional GANs, and produces results which are of comparable quality. We hope that this research encourages others to look into different ways of interpreting and optimizing distance metrics.

## 2. Related Work

GANs were originally proposed by Goodfellow *et al.* [8] in order to learn a sampling mechanism for complex data distributions. Intuitively, a generator $G_\theta(z)$, depending on parameters $\theta$, transforms perturbations $z$, obtained from a known distribution $\mathbb{P}_z$ over the latent space, into artificial samples. A discriminator $D_w(x)$, parameterized via $w$, compares the artificial samples to real world data points $x \in \mathcal{X}$ which we subsume in the dataset $\mathcal{D} = \{x\}$. We assume the data $\mathcal{D}$ to arise from an unknown data distribution $\mathbb{P}_d$ defined on a compact space $\mathcal{X}$. To compare data and artificial samples, the discriminator performs binary classification into "real" or "fake" by minimizing the negative log-likelihood, *i.e.*, $-\log D_w$ on real data points and $-\log(1 - D_w)$ on artificial samples, while the generator tries to make this minimization as hard as possible. Taken together, GANs address the following minimax program:

$$\max_\theta \min_w \quad \mathbb{E}_{x \sim \mathbb{P}_d}[-\log D_w(x)] \qquad (1)$$
$$+ \mathbb{E}_{z \sim \mathbb{P}_z}[-\log(1 - D_w(G_\theta(z)))].$$

For computational reasons, both expectations are evaluated empirically using samples. Impressive performance was demonstrated using this framework which also spurred a significant amount of work addressing possible improvements. In the following we discuss some of the issues that have been addressed to some degree in the past.

**Training Instability:** Training of GANs, *i.e.*, optimization of the program given in Eq. (1), is unstable in general, *e.g.*, well trained discriminators may suppress the training of generators. To address this issue, careful tuning of the number of generator updates after every discriminator update has been suggested. However, efforts like these are specific to tasks and hardly generalize. To understand this instability, Arjovsky *et al.* [1] showed that under the optimal discriminator, the training objective for the generator is equivalent to the inverted Kullback-Leibler divergence minus two times the Jensen-Shannon divergence between the data distribution $\mathbb{P}_d$ and the transformed sample distribution, *i.e.*, $G_\theta(\mathbb{P}_z)$. The negative Jensen-Shannon divergences term in the cost function pushes $\mathbb{P}_d$ and $G_\theta(\mathbb{P}_z)$ apart, contradicting the inverted Kullback-Leibler divergence's efforts to draw them closer.

**Mode Dropping:** Mode dropping refers to the phenomenon that generated samples lack diversity. For example, a generator for MNIST digits may suffer from the problem of "mode dropping" if it only generates a few of the ten digits. This problem has been observed when training GANs, especially in their "$-\log D$" incarnation of [31]. Again, Arjovsky *et al.* [1] provided some theoretical justification to this problem, arguing that the inverted Kullback-Leibler divergence is extremely benevolent to mode dropping but extremely harsh to novel samples.

It was shown in [2], that the aforementioned problems can be addressed by replacing the Jensen-Shannon divergence optimized in the original GAN framework with the Wasserstein-1 distance, also known as the Earth mover's distance. More specifically, the Wasserstein-p distance between the unknown data distribution $\mathbb{P}_d$ and the transformed latent distribution $G_\theta(\mathbb{P}_z)$, which are both defined on a compact data space $\mathcal{X}$, is given by

$$W_p(\mathbb{P}_d, G_\theta(\mathbb{P}_z)) = \inf_{\gamma \in \Pi(\mathbb{P}_d, G_\theta(\mathbb{P}_z))} (\mathbb{E}_{(x,y) \sim \gamma}[\|x - y\|^p])^{\frac{1}{p}}, \qquad (2)$$

where $\Pi(\mathbb{P}_d, G_\theta(\mathbb{P}_z))$ denotes all joint distributions that have marginals $\mathbb{P}_d$ and $G_\theta(\mathbb{P}_z)$. Computing the infimum in Eq. (2) is hard, partly because the data distribution $\mathbb{P}_d$ is not known. Therefore, it was proposed [2] to employ the Kantorovich-Rubinstein duality to Wasserstein-1 distance [34], which yields

$$W(\mathbb{P}_d, G_\theta(\mathbb{P}_z)) = \sup_{\|f\|_L \leq 1} \mathbb{E}_{x \sim \mathbb{P}_d}[f(x)] - \mathbb{E}_{z \sim \mathbb{P}_z}[f(G_\theta(z))], \qquad (3)$$

where the supremum is over all 1-Lipschitz functions $f : \mathcal{X} \to \mathbb{R}$. To approximate the maximization in Eq. (3), [2] proposed to train a neural network $f_w$ parametrized by weights $w \in \mathcal{W}$, which are clipped to ensure that $w$ lies in a compact space $\mathcal{W}$, enforcing $f_w$ to be $K$-Lipschitz for some $K$. Combined, the resulting Wasserstein GAN program

reads

$$\min_{\theta} \max_{w \in \mathcal{W}} \mathbb{E}_{x \sim \mathbb{P}_d}[f_w(x)] - \mathbb{E}_{z \sim \mathbb{P}_z}[f_w(G_\theta(z))]. \quad (4)$$

How to impose the Lipschitz constraint on the discriminator is still an open problem. Gradient clipping, illustrated in Eq. (4), was found to converge slowly and to have high variance. Gulrajani *et al.* [9] proposed a different method which restricts the norm of the gradient of the discriminator. This showed improvements over the original Wasserstein GAN, allowing for easier generalization of the method. However, since the Wasserstein GAN utilizes the discriminator to estimate the distance between the two distributions, the correctness of the estimate depends fundamentally on how well the discriminator has been trained. If the discriminator is not trained enough, the signal might completely mislead the generator. This is solved by training the discriminator several times before a single generator update. Furthermore, before the first generator update, the discriminator needs to be trained for a significant time to ensure progress, which adds computation cost and remains empirically motivated.

## 3. Approach

Following Arjovsky *et al.* [2], we also consider the Wasserstein distance to model distributions. But motivated by stability arguments and in contrast to using the Kantorovich-Rubinstein duality, we pursue *an approach that estimates the Wasserstein distance directly from samples*. This is based on random projections which will lead to the "sliced Wasserstein distance." Moreover, just like for the original GAN formulation given in Eq. (1), usage of the Kantorovich-Rubinstein duality, as outlined in Eq. (4) yields a saddle-point problem. However, saddle-point problems are generally hard to optimize, particularly if they are neither convex nor concave in any of the directions. Instead, our proposed *formulation searches for a global minimizer*. In addition, for the first time for GAN training, we are able to *show estimates for expected accuracy*.

To describe our approach, we first consider the Wasserstein distance between two datasets containing real data points $x \in \mathcal{D} \subseteq \mathcal{X}$, and artificially generated samples $\hat{x} = G_\theta(z) \in \mathcal{F} \subseteq \mathcal{X}$, which we subsume in the set of "fake" samples $\mathcal{F}$.

For notational simplicity only, we describe our proposed approach without introducing the notion of mini-batches. We however emphasize that mini-batches can be used in a straightforward manner.

Note that the quadratic Wasserstein distance $W_2^2(\mathcal{D}, \mathcal{F})$ between two sets of data points $\mathcal{D}$ and $\mathcal{F}$ is equivalently defined as

$$W_2^2(\mathcal{D}, \mathcal{F}) = \frac{1}{|\mathcal{F}|} \min_{\sigma \in \Sigma_{|\mathcal{F}|}} \sum_{i=1}^{|\mathcal{F}|} \|\mathcal{D}_{\sigma(i)} - \mathcal{F}_i\|_2^2, \quad (5)$$

where $\Sigma_{|\mathcal{F}|}$ is the set of all permutations of $|\mathcal{F}|$ elements. We use the subscript notation $\mathcal{D}_i$ and $\mathcal{F}_i$ to refer to the $i$-th sample in the dataset. Intuitively, the distance defined in Eq. (5) searches for a one-to-one assignment, *i.e.*, a bijective mapping of a "fake" sample $\mathcal{F}_i$ to a unique real data point $\mathcal{D}_{\sigma(i)}$ with index $\sigma(i)$ such that the squared difference accumulated over the entire dataset is minimal. Note that this assumes $|\mathcal{F}| = |\mathcal{D}|$, which is generally not a severe restriction, particularly when considering the fact that we generate the set of fake data $\mathcal{F}$.

To facilitate the computation of the distance defined in Eq. (5), the search for the optimal permutation $\sigma^*$ is reformulated as an integer linear program over the space of doubly stochastic matrices $M$ with integral entries, *i.e.*, matrices where both rows and columns sum to one:

$$W_2^2(\mathcal{D}, \mathcal{F}) = \frac{1}{|\mathcal{F}|} \min_M \sum_{i=1}^{|\mathcal{F}|} \sum_{j=1}^{|\mathcal{D}|} M_{i,j} \|\mathcal{D}_j - \mathcal{F}_i\|_2^2 \quad (6)$$

s.t. $M$ integral, doubly stochastic.

The matrix $M$ is also referred to as a permutation matrix and the task is known as a (linear) assignment problem as the cost function is linear in the entries of the argument $M$. Importantly, despite integrality constraints, a globally optimal solution for this program can be found with a linear programming solver because the constraint matrix of the program provided in Eq. (6) is totally unimodular [7]. Hence we can drop the integrality constraints while still obtaining an integral solution.

Note that this formulation is conceptually similar to the definition of the Wasserstein distance provided in Eq. (2). Although problems of this form can be solved with standard linear programming algorithms, dedicated methods are more suitable and achieve computational complexities of $O(|\mathcal{F}|^{2.5} \log(|\mathcal{F}|))$ [7]. Despite the availability of dedicated solvers for problems of the form given in Eq. (6), we found their complexity to be prohibitive for usage in the inner loop of a learning algorithm.

To address this issue we note that the 1-dimensional case, *i.e.*, the case where $x \in \mathbb{R}$ and $\hat{x} = G_\theta(z) \in \mathbb{R}$, has a more elegant solution. Specifically, let $\sigma_D$ and $\sigma_F$ be the permutations such that

$$\mathcal{D}_{\sigma_D(i)} \leq \mathcal{D}_{\sigma_D(i+1)}, \quad \forall i \in \{1 \leq i < N\}, \quad (7)$$

$$\mathcal{F}_{\sigma_F(i)} \leq \mathcal{F}_{\sigma_F(i+1)}, \quad \forall i \in \{1 \leq i < N\}. \quad (8)$$

Note that those permutations are easily obtained by sorting real data and artificial samples according to their value, which is possible in the 1-dimensional case. Given those permutations, the optimal $\sigma^*$ for the Wasserstein distance

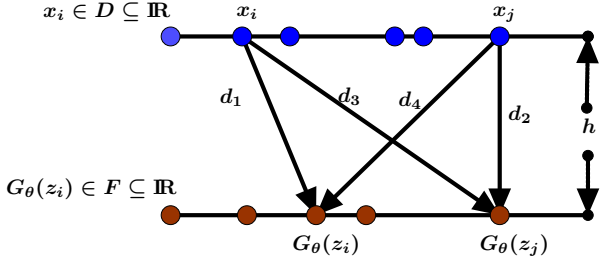

Figure 2. $\|x_i - G_\theta(z_i)\|_2^2 + \|x_j - G_\theta(z_j)\|_2^2 = (d_1^2 - h^2) + (d_2^2 - h^2) < (d_3^2 - h^2) + (d_4^2 - h^2) = \|x_i - G_\theta(z_j)\|_2^2 + \|x_j - G_\theta(z_i)\|_2^2$. Thus, the minimum in Eq. (5) is achieved when there is no line crossing, *i.e.*, samples monotonically arranged.

defined in Eq. (5) is simply

$$\sigma^* = \sigma_D \sigma_F^{-1}, \quad i.e., \tag{9}$$

$$W_2^2(\mathcal{D}, \mathcal{F}) = \frac{1}{|\mathcal{F}|} \sum_{i=1}^{|\mathcal{D}|} \|\mathcal{D}_{\sigma_D(i)} - \mathcal{F}_{\sigma_F(i)}\|_2^2. \tag{10}$$

Intuitively, the permutation $\sigma^*$ assigns the "fake" sample $\mathcal{F}_{\sigma_F(i)}$ to the real data point $\mathcal{D}_{\sigma_D(i)}$. To see that this is indeed the optimal assignment, let's consider Fig. 2 more carefully. The assignment is optimal if the data point with smallest value is assigned to the "fake" sample with smallest value. Any other assignment would result in crossing pairings which can be minimized further by disentangling the corresponding points. In practice, for 1-dimensional datasets (of identical size) we sort both $\mathcal{D}$ and $\mathcal{F}$ in $O(|\mathcal{F}| \log |\mathcal{F}|)$ time to find the correspondences and therefore the optimal permutation $\sigma^*$.

However, machine learning datasets of interest are rarely 1-dimensional. Therefore, in the following, we extend the 1-dimensional special case to an alternative metric. The employed technique is based on random projections of the high-dimensional datasets onto a variety of 1-dimensional subspaces. Formally, we project the datapoints and artificial examples onto 1-dimensional spaces by integrating over all possible directions $\omega \in \Omega$ on the unit sphere $\Omega$:

$$\tilde{W}_2^2(\mathcal{D}, \mathcal{F}) = \int_{\omega \in \Omega} W_2^2(\mathcal{D}^\omega, \mathcal{F}^\omega) d\omega. \tag{11}$$

Hereby the sets $\mathcal{D}^\omega = \{\omega^\top \mathcal{D}_i\}_{i=1}^{|\mathcal{D}|}$ and $\mathcal{F}^\omega = \{\omega^\top \mathcal{F}_i\}_{i=1}^{|\mathcal{F}|}$ contain 1-dimensional projections of the datapoints $\mathcal{D}_i$ and $\mathcal{F}_i$ onto the direction $\omega$. $\tilde{W}_2(\mathcal{D}, \mathcal{F})$ is also known as the "sliced Wasserstein distance" [6]. Kolouri *et al.* [15] have shown that the sliced Wasserstein distance satisfies the properties of non-negativity, identity of indiscernibles, symmetry, and subadditivity. Hence, it is a true metric.

In practice, we approximate the sliced Wasserstein distance between the distributions by using random samples and

---

**Algorithm 1:** Training the Sliced Wasserstein Generator

**Given** : Parameters $\theta$, sample size $n$, number of projections $m$, learning rate $\alpha$

1 **while** $\theta$ *not converged* **do**
2      Sample data $\{\mathcal{D}_i\}_{i=1}^n \sim \mathbb{P}_x$, noise $\{z_i\}_{i=1}^n \sim \mathbb{P}_z$;
3      $\{\mathcal{F}_i\}_{i=1}^n \leftarrow \{G_\theta(z_i)\}_{i=1}^n$;
4      compute **sliced Wasserstein Distance** $(\mathcal{D}, \mathcal{F})$
5          Init loss $L \leftarrow 0$;
6          Sample random projection directions $\Omega = \{\omega_{1:m}\}$;
7          **for** *each* $\omega \in \Omega$ **do**
8              $\mathcal{D}^\omega \leftarrow \{\omega^T D_i\}_{i=1}^n, \mathcal{F}^\omega \leftarrow \{\omega^T F_i\}_{i=1}^n$;
9              $\mathcal{D}_\sigma^\omega \leftarrow$ sorted $\mathcal{D}^\omega$, $\mathcal{F}_\sigma^\omega \leftarrow$ sorted $\mathcal{F}^\omega$;
10            $L \leftarrow L + \frac{1}{n}\|\mathcal{D}_\sigma^\omega - \mathcal{F}_\sigma^\omega\|^2$;
11          **end**
12          return $\frac{L}{m}$;
13      $\theta \leftarrow \theta - \alpha \nabla_\theta L$;
14 **end**

---

replacing the integration over $\Omega$ with a summation over a randomly chosen set of unit vectors $\hat{\Omega}$. It is now straightforward to formulate the optimization as

$$\min_\theta \frac{1}{|\hat{\Omega}|} \sum_{\omega \in \hat{\Omega}} W_2^2(\mathcal{D}^\omega, \mathcal{F}^\omega(\theta)), \tag{12}$$

when using the sliced Wasserstein distance metric. Hereby we made the dependence of the "fake" samples $\mathcal{F}(\theta)$ and their respective projections $\mathcal{F}^\omega(\theta)$ on the generator parameters $\theta$ explicit. Note that we obtain a single minimization as opposed to a saddle-point formulation.

We summarize the proposed approach in Alg. 1. In every iteration, we sample random directions (*e.g.*, from a Gaussian distribution). We then draw a set of samples from the true and fake distributions. Afterwards we project the distributions along each random direction, and compute the Wasserstein distance between the projected distributions. The sliced Wasserstein distance between the true and the fake distributions is computed as the average Wasserstein distance along all the projections. Gradients for the parameters of the deep net are computed by differentiating this loss, and any variant of stochastic gradient descent can be used to perform the parameter updates.

We also want to mention that computation of $W_2(\mathcal{D}^\omega, \mathcal{F}^\omega(\theta))$ requires a sorting algorithm, which increases computational complexity compared to optimizing GANs and GAN variants. This increase is slightly alleviated by the fact that our proposed technique does not need a discriminator. All in all, we found the generator updates of our approach to be slower by a factor varying from 1.5 to 2 in our experiments when $|\hat{\Omega}| \leq 10,000$ and when the sample size is less or equal to 256. More details are provided in Sec. 4. This may be time well spent when also considering the improved stability that we demonstrate in Sec. 4.

## 3.1. Training objective as an upper bound

In the following we provide a more formal treatment of the described approach and show that by training on the objective given in Eq. (12), we are, in fact, optimizing an upper bound on the sliced Wasserstein distance between the true distribution and the generated distribution.

Let $\mathbb{P}_f = G_\theta(\mathbb{P}_z)$ be the distribution induced by the generator. Our goal is to learn the data distribution $\mathbb{P}_d$. If $\hat{\mathbb{P}}_d, \hat{\mathbb{P}}_f$ are random empirical measures of $\mathbb{P}_d, \mathbb{P}_f$, then our optimization problem can be formulated as

$$\min_{\mathbb{P}_f} \mathbb{E}[\tilde{W}_2^2(\hat{\mathbb{P}}_d, \hat{\mathbb{P}}_f)]. \qquad (13)$$

However, we are concerned about $\tilde{W}_2^2(\mathbb{P}_d, \mathbb{P}_f)$. This is related to the program given in Eq. (13) in the following manner.

**Claim 1** *Let $\mathbb{P}_d$ and $\mathbb{P}_f$ be two distributions. Suppose that $\hat{\mathbb{P}}_d$ and $\hat{\mathbb{P}}_f$ are empirical measures of $\mathbb{P}_d$ and $\mathbb{P}_f$, induced by random sets (of $n$ i.i.d samples) $\mathcal{D}$ and $\mathcal{F}$. Then*

$$\tilde{W}_2^2(\mathbb{P}_d, \mathbb{P}_f) \leq 16 \mathbb{E}[\tilde{W}_2(\hat{\mathbb{P}}_d, \hat{\mathbb{P}}_f)]. \qquad (14)$$

**Proof:** See Appendix A.

Following the proof of Claim 1, we can guarantee the following bound for the generated distribution that solves our training objective.

**Corollary 1** *Let $\mathbb{P}_d$ and $\mathbb{P}_f$ be two distributions. Suppose that $\hat{\mathbb{P}}_d$ and $\hat{\mathbb{P}}_f$ are ($n$-sample) empirical measures of $\mathbb{P}_d$ and $\mathbb{P}_f$, and let $\hat{\mathbb{P}}'_d$ be an independent copy of $\hat{\mathbb{P}}_d$. For $\mathbb{P}_f^*$ defined by $\mathbb{P}_f^* = \operatorname{argmin}_{\mathbb{P}_f} \mathbb{E}[\tilde{W}_2^2(\hat{\mathbb{P}}_d, \hat{\mathbb{P}}_f)]$, the following holds:*

$$\tilde{W}_2(\mathbb{P}_d, \mathbb{P}_f^*) \leq 14 \mathbb{E}[\tilde{W}_2(\hat{\mathbb{P}}_d, \hat{\mathbb{P}}'_d)]. \qquad (15)$$

**Proof:** See Appendix B.

Corollary 1 tells us that, as $\mathbb{E}[\tilde{W}_2^2(\hat{\mathbb{P}}_d, \hat{\mathbb{P}}'_d)] \to 0$, our bound gets tighter and therefore we should be able to learn a better solution. We investigate how $\mathbb{E}[\tilde{W}_2^2(\hat{\mathbb{P}}_d, \hat{\mathbb{P}}'_d)]$ behaves empirically for different datasets with the number of samples used in Sec. 4.1.

## 3.2. Scaling to high dimensional distributions

By minimizing the sliced Wasserstein distance between the distributions $\mathbb{P}_d$ and $G_\theta(\mathbb{P}_z)$ over a finite set of directions, we are essentially matching marginals of $\mathbb{P}_d$ and $G_\theta(\mathbb{P}_z)$ along those directions. For faster convergence it is, therefore, better to use projections along which the distributions are most dissimilar. Since we are randomly sampling projections in a high dimensional space, it is unlikely that all projections sampled will have useful information, especially as training progresses.

| Dataset | Size | Approx #examples |
|---|---|---|
| MNIST | 28x28x1 | 50,000 |
| CIFAR-10 | 32x32x3 | 50,000 |
| TFD | 48x48x1 | 100,000 |
| LSUN Bedrooms | 64x64x3 | 200,000 |
| CelebA | 64x64x3 | 200,000 |

Table 1. Datasets used in various experiments

In theory this can be addressed by methods such as linear discriminant analysis, but they are expensive. Instead we choose to use a discriminator, much like those in GANs, to provide 'good' projections. Put simply, a neural network based discriminator tries to map the real and fake samples into a space where it is easy to tell them apart. Any projection in this space will have significantly more information, since the two classes of samples are better separated in this space. Suppose the output of some intermediate layer of the neural network can be expressed as the function $f_{\theta'}$, while the overall discriminator is the function $f'_{\theta'}$. Then, instead of matching the distributions of $\mathbb{P}_d$ and $G_\theta(\mathbb{P}_z)$, we train our generator to match the distributions of $f_{\theta'}(\mathbb{P}_d)$ and $f_{\theta'}(G_\theta(\mathbb{P}_z))$. The two objectives, which are optimized independently of each other are:

$$\min_\theta \frac{1}{|\hat{\Omega}|} \sum_{\omega \in \hat{\Omega}} W_2^2(f_{\theta'}(\mathcal{D})^\omega, f_{\theta'}(\mathcal{F})^\omega(\theta)),$$

$$\min_{\theta'} \mathbb{E}[-\log(f'_{\theta'}(\mathcal{D}))] + \mathbb{E}[-\log(1 - f'_{\theta'}(\mathcal{F}))],$$

for $\mathcal{D} \sim \mathbb{P}_d, \mathcal{F} \sim G_\theta(\mathbb{P}_z)$. We find that this heuristic is robust to different discriminator architectures. This is demonstrated empirically in Sec. 4.4.

## 4. Experimental Results

In this section, we present results to (1) compare the training of a generator with our method (henceforth called the sliced Wasserstein Generator, or SWG) to other generative models, and to (2) show how our method is stable across different architectures of the generator and discriminator. We use several datasets for our experiments. These are summarized in Tab. 1. Baselines are the GAN in its "-log D" incarnation [31], and the Wasserstein GAN (with gradient penalty) from [9].

### 4.1. Effect of sample size

In our first experiment we investigate the upper bound of Corollary 1. We compute empirically for different datasets,

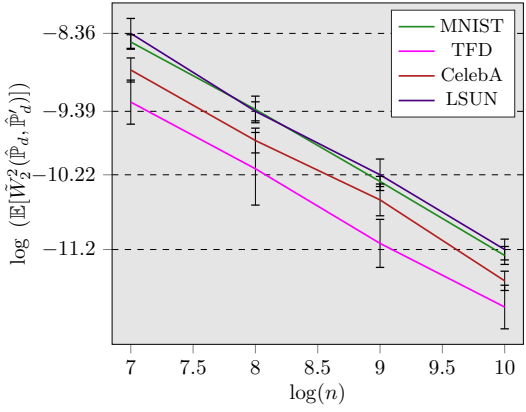

Figure 3. Limited sample estimate of the sliced Wasserstein distance as a function of the sample size.

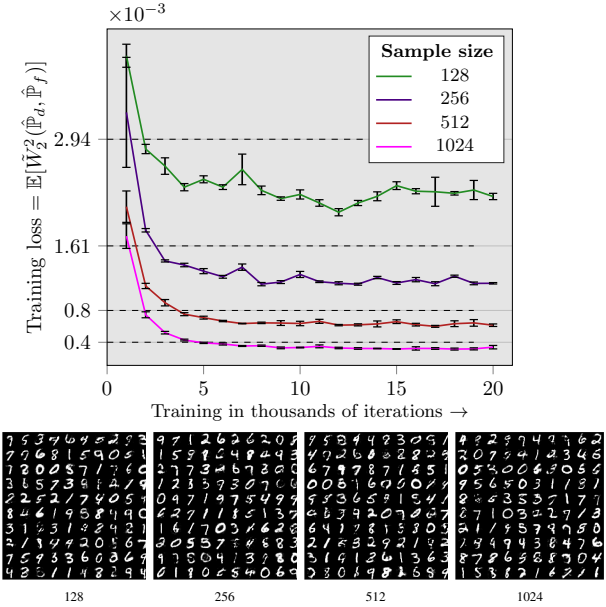

Figure 4. Training with different sample sizes on MNIST. The dashed lines denote $\mathbb{E}[\tilde{W}_2^2(\hat{\mathbb{P}}_d, \hat{\mathbb{P}}'_d)]$.

and show in Fig. 3, how $\mathbb{E}[\tilde{W}_2^2(\hat{\mathbb{P}}_d, \hat{\mathbb{P}}'_d)]$ decreases with the number of samples used for estimation. To obtain this quantity we take two sets of $n$ samples, each from the data distribution $\mathbb{P}_d$. We then compute the sliced Wasserstein distance between those sets in the manner described in Alg. 1. We observe that $\mathbb{E}[\tilde{W}_2^2(\hat{\mathbb{P}}_d, \hat{\mathbb{P}}'_d)]$ decreases roughly via $O(n^{-1})$. Using Corollary 1, this implies that $\tilde{W}_2^2(\mathbb{P}_d, \mathbb{P}_f^*)$ decreases in $O(n^{-1})$ for the optimal solution $\mathbb{P}_f^*$.

To test the quality of this loss estimate, we train a fully connected deep net based generator on the sliced Wasserstein distance with different sample sizes for the MNIST dataset. Each configuration was trained 5 times with randomly set seeds, and the averages with error bars are presented in Fig. 4. During training, at every iteration, gradients are computed using 10,000 random projections. We emphasize the small

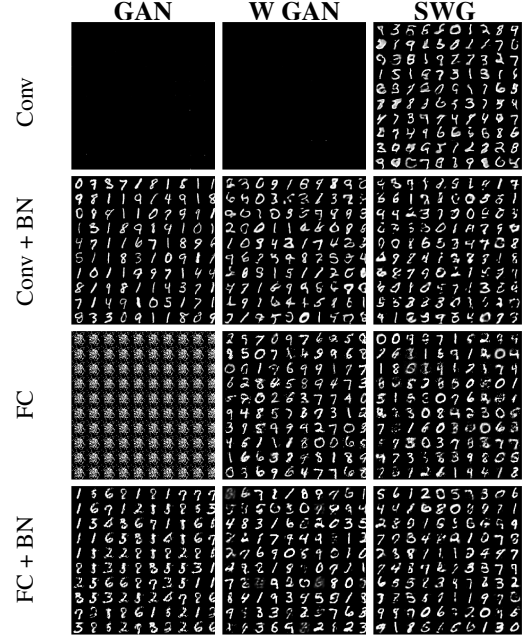

Figure 5. MNIST samples after 40k training iterations for different generator configurations. Batch size $= 250$, Learning rate $= 0.0005$, Adam optimizer

error bars which highlight the stability of the proposed approach.

The generator is able to produce good images in all four cases. This shows that, in practice, a set of as few as 128 samples is good enough for simple distributions. The generator is able to beat $\mathbb{E}[\tilde{W}_2^2(\hat{\mathbb{P}}_d, \hat{\mathbb{P}}'_d)]$ (dashed black line) on the loss, indicating that it has probably converged in all cases. As the number of samples increases, we see this bound getting tighter.

### 4.2. Stability of Training

To demonstrate the stability of the proposed approach, four different generator architectures are trained with our method as well as the two aforementioned baselines using exactly the same set of hyperparameters. One generator is composed of fully connected layers while the other is composed of convolutional and deconvolutional layers. For each generator we assess its performance when using and when not using batch normalization [12]. The architectures are described in more detail in Appendix D. For this experiment, only the GAN and Wasserstein GAN use a discriminator, while our approach relies on random projections instead. Further note that these architectures are arbitrarily chosen, and this comparison is only intended to show how the training stability compares across different methods, as well as how the sliced Wasserstein loss correlates with the generated samples. This is not to compare the best possible samples from different training methods.

Samples obtained from the resulting generator are visu-alized in Fig. 5. We observe that only the SWG is able to produce meaningful samples in every configuration. Surprisingly, even the Wasserstein GAN fails in one of the configurations. The SWG is more robust in this setting than the Wasserstein GAN, while needing less computation since the Wasserstein GAN requires multiple discriminator updates per generator update. This is more expensive than the extra computation required for sorting in SWG.

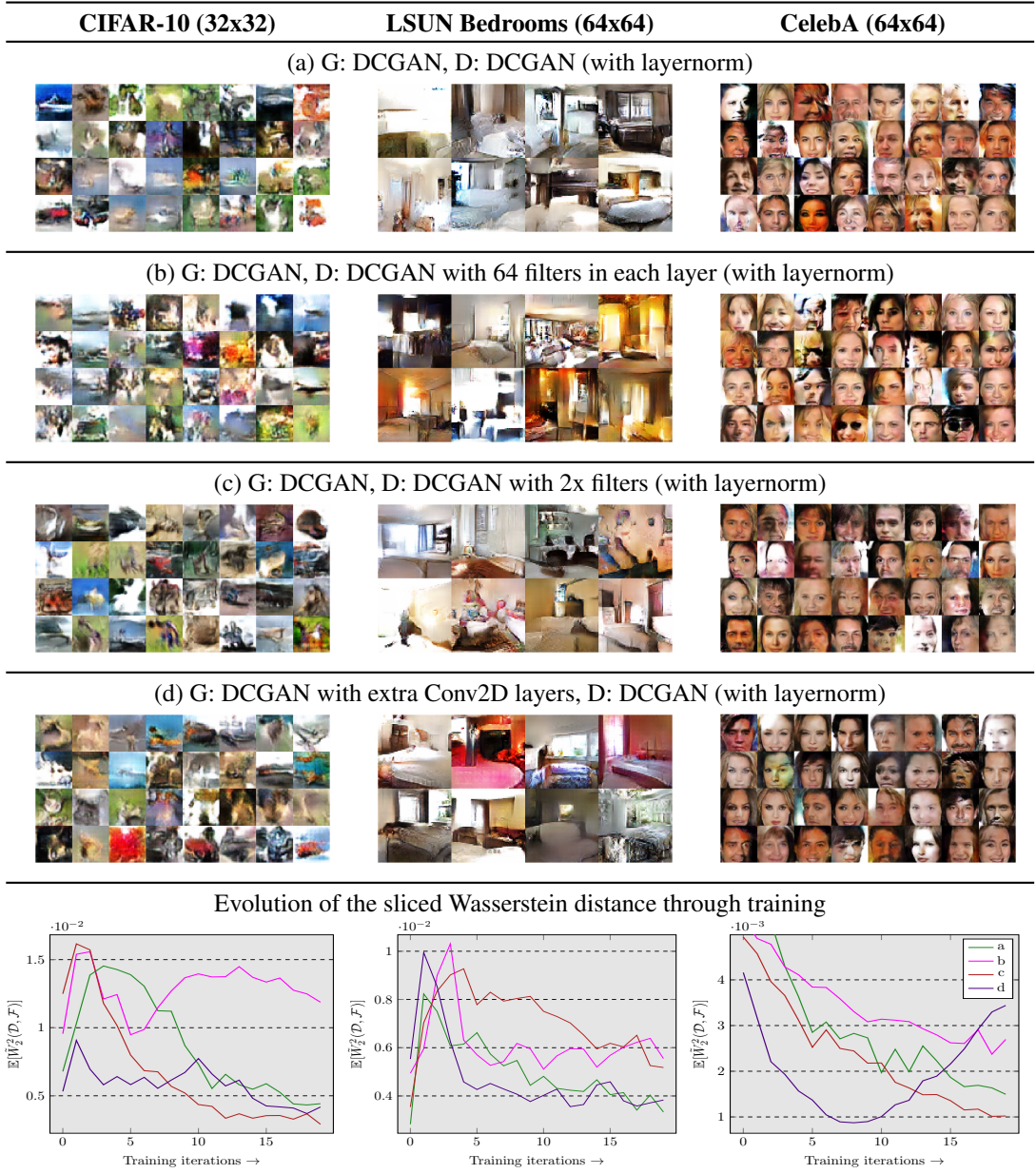

Table 2. The SWG succeeds in training different architectures (a through d) on different datasets with same hyperparameters. Samples collected after 20 epochs of training with batch size = 64, learning rate = 0.0001, Adam optimizer

To analyze this result more carefully, in Fig. 6, we show how two metrics, namely the symmetrized KL divergence, and the sliced Wasserstein distance, evolve over the training iterations. These results are averaged over 5 runs, and plotted with error bars that represent the standard deviation. The KL divergence is computed using the ITE toolbox from [33]. The sliced Wasserstein distance is calculated as the mean computed with a fixed set of 100,000 projection directions. We show this for the convolutional generator with and without batch norm. SWG is extremely stable, with the KL divergence improving through training. The Wasserstein GAN shows very high variance. We do not speculate here about the cause, but merely state the observation that the training objective is not very stable. The GAN diverges through training. This is a known behavior of GANs.

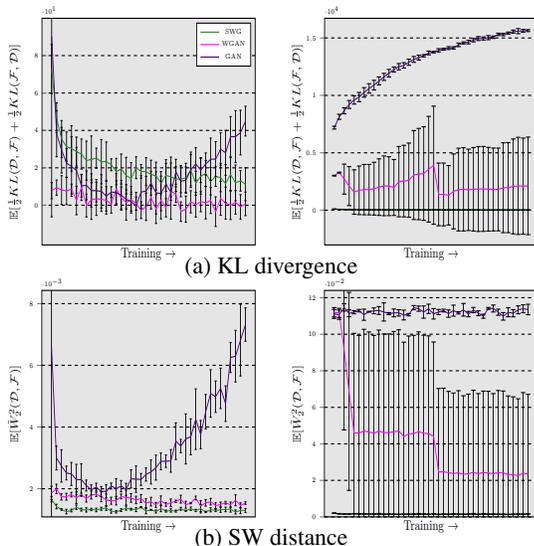

Figure 6. Training progress on MNIST for the Conv + BN and Conv generators. Estimated using 500 samples each from both distributions.

### 4.3. Effectiveness of the sliced Wasserstein distance

From Fig. 5 and Fig. 6, we observe how the sliced Wasserstein distance correlates with sample outputs. The GAN performs poorly on this metric even though it produces good images. By inspection of the resulting samples, it is clear that the GAN suffers from mode collapse around the digit 1. Although the images are sharp, they lack sample diversity. Because of this, the generated distribution is at a greater sliced Wasserstein distance from the true distribution.

The sliced Wasserstein distance appears to be less harsh on sample quality, and, because of this, the WGAN and the SWG are able to achieve a better performance. Since the WGAN and the SWG are optimizing different interpretations of the same distance, this is perhaps not surprising.

The SWG produces good, diverse samples, and is able to perform best on this distance. Interestingly, the divergent behavior of the GAN is observable early on when using the sliced Wasserstein distance. Our experiments indicate that the sliced Wasserstein distance is a good measure for distance between two distributions, taking into account both, the sample quality, and the sample diversity.

### 4.4. Scaling to high dimensional distributions

In this section and in Tab. 2, we present results on the CIFAR-10, LSUN Bedrooms, and CelebA datasets (columns in Tab. 2 using the training method described in Sec. 3.2. Along with a generator, we also use a discriminator and we match distributions in the penultimate layer of the discriminator. To show the robustness of our approach, we train with different architectures (rows in Tab. 2) while keeping the same hyperparameters across all experiments. The discriminator is trained once per generator update for these experiments. With a single default setting of hyperparameters, we succeed in training all architectures across all datasets.

| Projections | Batch Size | Time (s) |
|---|---|---|
| 5000 | 64 | 0.06 |
| 5000 | 256 | 0.146 |
| 10000 | 64 | 0.072 |
| 10000 | 256 | 0.17 |
| WGAN | 64 | 0.046 |
| WGAN | 256 | 0.13 |

Table 3. Comparison of time required for generator updates

The base architecture for both the generator and discriminator is the DCGAN [29]. Like [9] we use layer normalization [3] in the discriminator. We make modifications to this, for instance using twice as many filters in each layer of the discriminator, or using a constant 64 filters in every layer. We also test a deeper generator by adding 2 convolutional layers of stride 1 for one experiment.

We experiment with more discriminator training frequencies (*i.e.*, number of generator updates per discriminator update) and show results in Appendix C.

### 4.5. Training time comparison

We compared the time per iteration for a generator update in the SWG to a WGAN iteration. Both use Tensorflow v1.4 on a NVIDIA Tesla P-100 GPU. The results are summarized in Tab. 3. Due to sorting, SWG is slower by a factor of about 1.5 on the configurations tested. However, we do not require multiple discriminator updates per generator update, and therefore our approach is actually faster than the WGAN per generator update.

## 5. Conclusions

In this paper we proposed to use the sliced Wasserstein distance for generative modeling. We illustrated its efficacy on the MNIST dataset [18], the CIFAR-10 dataset [17], the CelebA dataset [21], and the LSUN dataset [35], and showed stable results that are competitive with existing techniques. Our implementation is publicly available[1].

**Acknowledgements:** This material is based upon work supported in part by the National Science Foundation under Grant No. 1718221. We thank NVIDIA for providing the GPUs used for this research.

---

[1]https://github.com/ishansd/swg

# Appendix

## A. Training objective as an upper bound

**Claim 1** *Let $\mathbb{P}_d$ and $\mathbb{P}_f$ be two distributions. Suppose that $\hat{\mathbb{P}}_d$ and $\hat{\mathbb{P}}_f$ are empirical measures of $\mathbb{P}_d$ and $\mathbb{P}_f$, induced by random sets (of $n$ i.i.d samples) $\mathcal{D}$ and $\mathcal{F}$. Then*

$$\tilde{W}_2^2(\mathbb{P}_d, \mathbb{P}_f) \leq 16 \mathbb{E}[\tilde{W}_2(\hat{\mathbb{P}}_d, \hat{\mathbb{P}}_f)]. \tag{16}$$

**Proof:** Using the triangle inequality for the sliced Wasserstein distance, we have

$$\tilde{W}_2^2(\mathbb{P}_d, \mathbb{P}_f) \leq 2\tilde{W}_2^2(\mathbb{P}_d, \hat{\mathbb{P}}_d) + 2\tilde{W}_2^2(\mathbb{P}_f, \hat{\mathbb{P}}_d). \tag{17}$$

Using it again, we get

$$\tilde{W}_2^2(\mathbb{P}_d, \mathbb{P}_f) \leq 2\tilde{W}_2^2(\mathbb{P}_d, \hat{\mathbb{P}}_d) + 4\tilde{W}_2^2(\mathbb{P}_f, \hat{\mathbb{P}}_f) + 4\tilde{W}_2^2(\hat{\mathbb{P}}_d, \hat{\mathbb{P}}_f). \tag{18}$$

In the following we find upper bounds for $\tilde{W}_2^2(\mathbb{P}_f, \hat{\mathbb{P}}_f)$ in terms of $\tilde{W}_2^2(\hat{\mathbb{P}}_d, \hat{\mathbb{P}}_f)$. In order to do this, we must deconstruct the sliced Wasserstein distance. By definition, we have

$$\tilde{W}_2^2(\mathbb{P}_f, \hat{\mathbb{P}}_f) = \int_{\omega \in \Omega} W_2^2(\mathbb{P}_f^\omega, \hat{\mathbb{P}}_f^\omega) d\omega. \tag{19}$$

Consider any one projection $\omega$. We have a 1-d distribution $\mathbb{P}_f^\omega$, and its empirical measure $\hat{\mathbb{P}}_f^\omega$. Using Theorem 4.3 in [5]:

$$\mathbb{E}[W_2^2(\mathbb{P}_f^\omega, \hat{\mathbb{P}}_f^\omega)] \leq \mathbb{E}[W_2^2(\hat{\mathbb{P}}_f^\omega, \hat{\mathbb{P}}_f'^\omega)], \tag{20}$$

where $\hat{\mathbb{P}}_f'^\omega$ is an independent copy of $\hat{\mathbb{P}}_f^\omega$.

To bound $\mathbb{E}[W_2^2(\hat{\mathbb{P}}_f^\omega, \hat{\mathbb{P}}_f'^\omega)]$ in Eq. (20), we first see how the expectsed Wasserstein distance between two 1-d empirical measures $\hat{\mathbb{P}}_d^\omega$ and $\hat{\mathbb{P}}_f^\omega$ can be written in terms of the sets of samples $\mathcal{D}^\omega$ and $\mathcal{F}^\omega$ that they represent (i.e. are induced by). Note that $\mathcal{D}^\omega$ and $\mathcal{F}^\omega$ are obtained by simply projecting a the sets $\mathcal{D}$ and $\mathcal{F}$ onto the direction $\omega$. If $\mathcal{D}_{\sigma_D(i)}^\omega$ and $\mathcal{F}_{\sigma_F(i)}^\omega$ denote the $i$-th smallest sample in $\mathcal{D}^\omega$ and $\mathcal{F}^\omega$,

$$\mathbb{E}[W_2^2(\hat{\mathbb{P}}_d^\omega, \hat{\mathbb{P}}_f^\omega)] = \frac{1}{n} \sum_{i=1}^n \mathbb{E}[(\mathcal{D}_{\sigma_D(i)}^\omega - \mathcal{F}_{\sigma_F(i)}^\omega)^2]. \tag{21}$$

$\mathcal{D}_{\sigma_D(i)}^\omega$ and $\mathcal{F}_{\sigma_F(i)}^\omega$ are infact the $n$ sample order statistics of $\mathbb{P}_d^\omega$ and $\mathbb{P}_f^\omega$. For $\hat{\mathbb{P}}_f^\omega$ and $\hat{\mathbb{P}}_f'^\omega$, we can write this as

$$\mathbb{E}[W_2^2(\hat{\mathbb{P}}_f^\omega, \hat{\mathbb{P}}_f'^\omega)] = \frac{2}{n} \sum_{i=1}^n Var[\mathcal{F}_{\sigma_F(i)}^\omega]. \tag{22}$$

The RHS of Eq. (21) can be decomposed as

$$\mathbb{E}[(\mathcal{D}_{\sigma_D(i)}^\omega - \mathcal{F}_{\sigma_F(i)}^\omega)^2]$$
$$= \mathbb{E}[(\mathcal{D}_{\sigma_D(i)}^\omega - \mathbb{E}[\mathcal{F}_{\sigma_F(i)}^\omega] + \mathbb{E}[\mathcal{F}_{\sigma_F(i)}^\omega] - \mathcal{F}_{\sigma_F(i)}^\omega)^2]$$
$$= Var[\mathcal{F}_{\sigma_F(i)}^\omega] + E[(\mathcal{D}_{\sigma_D(i)}^\omega - \mathbb{E}[\mathcal{F}_{\sigma_F(i)}^\omega])^2]$$
$$\geq Var[\mathcal{F}_{\sigma_F(i)}^\omega],$$

hence

$$\frac{1}{n} \sum_{i=1}^n Var[\mathcal{F}_{\sigma(i)}^\omega] \leq \frac{1}{n} \sum_{i=1}^n \mathbb{E}[(\mathcal{D}_{\sigma_D(i)}^\omega - \mathcal{F}_{\sigma_F(i)}^\omega)^2].$$

Combining this result with Eq. (21) and Eq. (22) yields

$$\mathbb{E}[W_2^2(\hat{\mathbb{P}}_f^\omega, \hat{\mathbb{P}}_f'^\omega)] \leq 2\mathbb{E}[W_2^2(\hat{\mathbb{P}}_d^\omega, \hat{\mathbb{P}}_f^\omega)],$$

which, when combined with Eq. (20), results in

$$\mathbb{E}[W_2^2(\mathbb{P}_f^\omega, \hat{\mathbb{P}}_f^\omega)] \leq 2\mathbb{E}[W_2^2(\hat{\mathbb{P}}_d^\omega, \hat{\mathbb{P}}_f^\omega)]. \tag{23}$$

Applying the expectation operator on Eq. (19) and using Eq. (23),

$$\begin{aligned}\mathbb{E}[\tilde{W}_2^2(\mathbb{P}_f, \hat{\mathbb{P}}_f)] &\leq 2\int_{\omega \in \Omega} \mathbb{E}[W_2^2(\hat{\mathbb{P}}_d^\omega, \hat{\mathbb{P}}_f^\omega)]d\omega \\ &= 2\mathbb{E}[\tilde{W}_2^2(\hat{\mathbb{P}}_d, \hat{\mathbb{P}}_f)].\end{aligned} \tag{24}$$

The same bound holds for $\mathbb{E}[\tilde{W}_2^2(\mathbb{P}_d, \hat{\mathbb{P}}_d)]$.

Substituting from Eq. (24) in Eq. (18) and applying the expectation operator, we get

$$\tilde{W}_2^2(\mathbb{P}_d, \mathbb{P}_f) \leq 16\mathbb{E}[\tilde{W}_2(\hat{\mathbb{P}}_d, \hat{\mathbb{P}}_f)], \tag{25}$$

which completes the proof. ∎

## B. Bounds for generated distribution

**Corollary 1** *Let $\mathbb{P}_d$ and $\mathbb{P}_f$ be two distributions. Suppose that $\hat{\mathbb{P}}_d$ and $\hat{\mathbb{P}}_f$ are (n-sample) empirical measures of $\mathbb{P}_d$ and $\mathbb{P}_f$, and let $\hat{\mathbb{P}}'_d$ be an independent copy of $\hat{\mathbb{P}}_d$. For $\mathbb{P}_f^*$ defined by $\mathbb{P}_f^* = \mathrm{argmin}_{\mathbb{P}_f} \mathbb{E}[\tilde{W}_2^2(\hat{\mathbb{P}}_d, \hat{\mathbb{P}}_f)]$, the following holds:*

$$\tilde{W}_2(\mathbb{P}_d, \mathbb{P}_f^*) \leq 14\mathbb{E}[\tilde{W}_2(\hat{\mathbb{P}}_d, \hat{\mathbb{P}}'_d)]. \tag{26}$$

**Proof:** This follows easily from Claim 1. Using Eq. (20), we can show that

$$\mathbb{E}[\tilde{W}_2^2(\mathbb{P}_d, \hat{\mathbb{P}}_d)] \leq \mathbb{E}[\tilde{W}_2^2(\hat{\mathbb{P}}_d, \hat{\mathbb{P}}'_d)], \tag{27}$$

and therefore we can rewrite (18) as:

$$\tilde{W}_2(\mathbb{P}_d, \mathbb{P}_f) \leq 2\mathbb{E}[\tilde{W}_2^2(\hat{\mathbb{P}}_d, \hat{\mathbb{P}}'_d)] + 12\mathbb{E}[\tilde{W}_2(\hat{\mathbb{P}}_d, \hat{\mathbb{P}}_f)]. \tag{28}$$

Since $\mathbb{P}_f^*$ minimizes $\mathbb{E}[\tilde{W}_2^2(\hat{\mathbb{P}}_d, \hat{\mathbb{P}}_f)]$ over all $\mathbb{P}_f$,

$$\mathbb{E}[\tilde{W}_2(\hat{\mathbb{P}}_d, \hat{\mathbb{P}}_f^*)] \leq \mathbb{E}[\tilde{W}_2(\hat{\mathbb{P}}_d, \hat{\mathbb{P}}'_d)]. \tag{29}$$

Therefore,

$$\tilde{W}_2(\mathbb{P}_d, \mathbb{P}_f^*) \leq 14\mathbb{E}[\tilde{W}_2(\hat{\mathbb{P}}_d, \hat{\mathbb{P}}'_d)]. \tag{30}$$

∎

## C. Discriminator update frequency experiments

We tested different discriminator update schemes (*i.e.*, number of generator updates per discriminator updates, and number of iterations of discriminator updates). In Tab. 4 we show samples after 40 epochs of training on the LSUN dataset with these different schemes for two discriminator configurations. The generator architecture for both is the DCGAN.

| Discriminator:**DCGAN** | **DCGAN with 64 filters in each layer** |

(a) 1 D update per G update, 1 iteration of training per D update

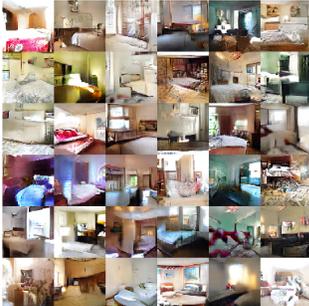 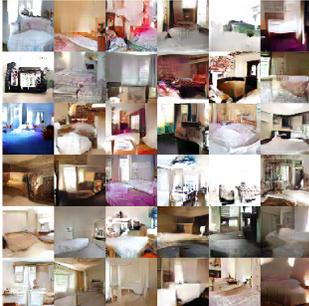

(b) 1 D update per G update, 5 iterations of training per D update

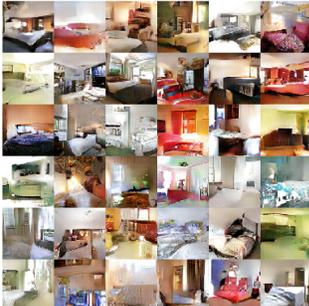 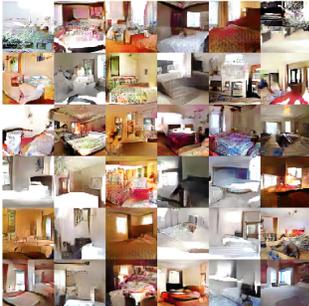

(c) 1 D update per 5 G updates, 1 iteration of training per D update

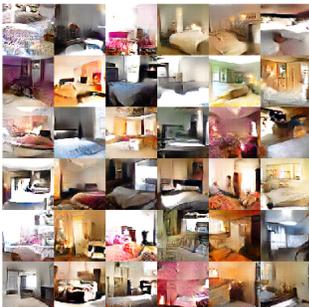 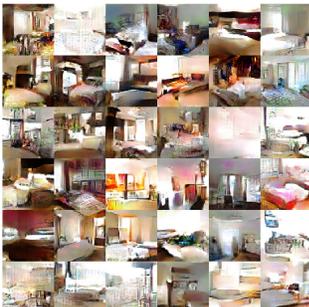

(d) 1 D update per 5 G updates, 5 iterations of training per D update

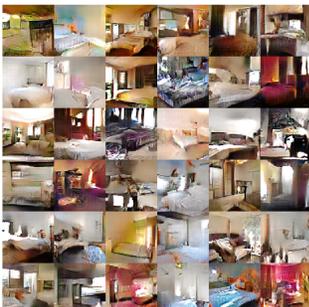 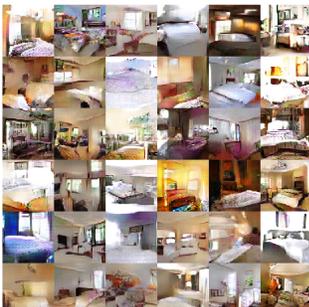

Table 4. The SWG is robust to different discriminator update schemes. Tested for two discriminator architectures (columns). Sample size = 64, learning rate = 0.0005, Adam optimizer, 40 epochs.

## D. Network architectures for experiments on MNIST

Here we summarize the different network architectures used for experiments with the MNIST dataset presented in Sec. 4.2.

| Generator (Fully Connected) | Generator (Conv & Deconv) | Discriminator |
|:---:|:---:|:---:|
| **output:** 784-d sample | **output:** 784-d sample | **output:** scalar |
| fc-784, sigmoid | conv2d-1-3-1, sigmoid | 2× fc-256, relu |
| 7× fc-512, relu | deconv2d-16-3-2, (bn), relu | **input:** 784-d sample |
| **input:** 32-d random noise | conv2d-32-3-1, (bn), relu | |
| | deconv2d-32-3-2, (bn), relu | |
| | conv2d-64-3-1, (bn), relu | |
| | deconv2d-64-3-2, (bn), relu | |
| | fc-1024 | |
| | **input:** 32-d random noise | |

Table 5. Generator and discriminator for MNIST. "fc-$n$" means applying a fully connected layer with $n$ output units. Both "conv2d-$c$-$k$-$s$" and "deconv2d-$c$-$k$-$s$" mean applying $c$ convolutional filters of size $k$ by $k$ with stride $s$ by $s$. "bn" means batch normalization.